\documentclass[11pt,a4paper]{article}
\usepackage{graphicx}

\title{\bf DISCRETE OPTIMIZATION\\ OF STATISTICAL SAMPLE
SIZES\\ IN SIMULATION BY USING\\ THE HIERARCHICAL\\ BOOTSTRAP METHOD
\thanks{We are very  thankful to Latvian Council of Science for
the Grant Nr.97.0798 within which the present investigation is worked out.}}

\author{A. Andronov and M. Fioshin\\[3mm]
Riga Technical university,\\
Riga, LV-1658, 1, Kalku Str., Latvia\\
{\it e-mail: andronow@rau.lv, mf@rau.lv}\\
}
\date{}
\begin{document}
\maketitle

\begin{abstract}
The Bootstrap method application in simulation supposes that value of
random variables are not generated during the simulation process but
extracted from available sample populations. In the case of
Hierarchical Bootstrap the function of interest is calculated recurrently
using the calculation tree. In the present paper we consider the
optimization of sample sizes in each vertex of the calculation tree.
The dynamic programming method is used for this aim. Proposed method
allows to decrease a variance of system characteristic estimators.
\\[0.3cm]
\hspace*{-0.5cm} {\it  Keywords:}
Bootstrap method, Simulation, Hierarchical Calculations, Variance Reduction
\end{abstract}

\setlength{\baselineskip}{20pt}
\section{INTRODUCTION}
Main problem of the mathematical statistics and simulation is connected with
insufficiency of primary statistical data. In this situation, the Bootstrap
method can be used successfully (Efron, Tibshirani 1993, 
Davison, Hinkley 1997). If a dependence between characteristics
of interest and input data is very composite and is described by numerical
algorithm then usually it applies a simulation. By this the probabilistic
distributions of input data are not estimated because the given primary
data has small size and such estimation gives a bias and big variance.
The Bootstrap method supposes that random variables are not generated
by a random number generator during simulation in accordance with the estimated distributions
but ones are extracted from given primary data at random.
Various problems of this approach were considered in previous papers
(Andronov et al. 1995, 1998).

We will consider the known function
$\phi$ of m independent continuos random variables $X_1, X_2, \ldots, X_m:
\phi (X_1, X_2, \ldots, X_m).$ It is assumed that distributions of random
variables $\{X_i\}$ are unknown, but the sample population
$H_i=\{X_{i1}, X_{i2}, \ldots, X_{in_i}\}$ is available for each
$X_i, i=\overline{1,m}$. Here $n_i$ is the size of the sample $H_i$.
The problem consists in estimation of the mathematical expectation
\begin{equation}
\label{thetaform}
\Theta=E\;\phi(X_1, X_2, \ldots, X_m).
\end{equation}

The Bootstrap method use supposes an organization of some realizations
of the values
$\phi(X_1, X_2, \ldots, X_m)$. In each realization the values of
arguments are extracted randomly from the corresponding sample populations
$H_1, H_2, \ldots, H_m$. Let $j_i(l)$ be a number of elements which were
extracted from the population $H_i$ in the $l$-th realization.
We denote ${\bf X}(l)=(X_{1,j_1(l)}, X_{2,j_2(l)}, \ldots, X_{m,j_m(l)})$
and name it the l-th subsample. The estimator $\Theta^*$ of the mathematical
expectation $\Theta$ is equal to an average value for all r realizations:
\begin{equation}
\label{thetaest}
\Theta^*=\frac{1}{r}\sum_{l=1}^r\phi({\bf X}(l)).
\end{equation}

Our main aim is to calculate and to minimize the variance of this estimator. The variance will
depend upon two factors: 1) a calculation method of the function $\phi$;
2) a formation mode of subsamples ${\bf X}(l)$. 

The next two Sections will
be dedicated to these questions. 
In Section 4 we will show how to decrease variance $D\;\Theta^*$ using
the dynamic programming method.

\section{THE HIERARCHICAL BOOTSTRAP METHOD}

We suppose that the function $\phi$ is calculated by a calculation tree.
A root of this tree corresponds to the computed function
$\phi=\phi_k$. It is the vertex number k. The vertices numbers $1, 2, \ldots, m$
correspond to the input variables $X_1, X_2, \ldots, X_m$. The rest vertices
are intermediate ones. They correspond to intermediate functions
$\phi_{m+1}, \phi_{m+2}, \ldots, \phi_{k-1}$ (see Fig.1).

\begin{figure}[h]
\includegraphics{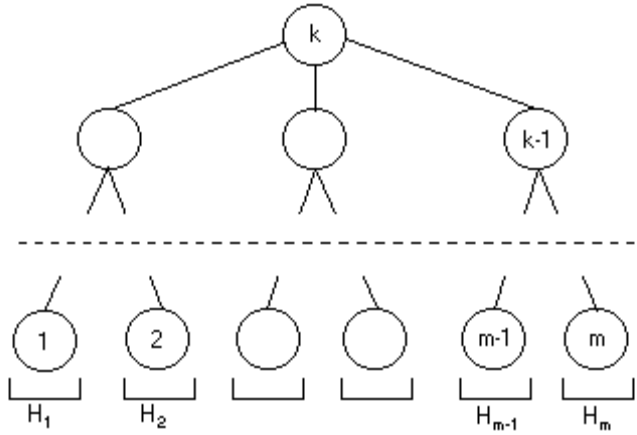}
\bigskip
\bigskip
\bigskip
\caption{Calculation tree}
\label{f2}
\end{figure}

Only one arc $y_v$ comes out from each vertex $v$, $v=1,2,\ldots,k$. It
corresponds to a value of the function $\phi_v$. We suppose that
$y_v=X_v, v=1,2,\ldots,m; y_k=\phi=\phi_k$.

We denote $I_v$ as a set of vertices from which arcs come into the vertex
$v$, and $J_v$ as a corresponding set of variables (arcs):
$i\in I_v\Leftrightarrow y_i\in J_v$. It is clear that $I_v=\oslash$ for
$v=1,2,\ldots,m$; $y_v=f_v(J_v)$, $v=m+1, m+2, \ldots, k$. We suppose that
a numbering of the vertices is correct: if $v<w$ than $w\notin I_v$.

Now, function value can be calculated by the sweep method. At the
beginning, we extract separate elements $X_{1j_1(1)}, X_{2j_2(1)}, \ldots,
X_{mj_m(1)}$ from populations $H_1, H_2, \ldots, H_m$, then calculate the
function values $\phi_{m+1}, \phi_{m+2}, \ldots$, $\phi_k=\phi$ successively.
After r such runs the estimator $\Theta^*$ is computed according to the
formula (\ref{thetaest}). An analysis of this method was developed in
the previous papers of authors.

The Hierarchical Bootstrap method is based on the wave algorithm. Here
all values of the function $\phi_v$ for each vertex $v=m+1,m+2,\ldots,
k$ should be calculated all at once. They form the set $H_v=\{Y_{v1},Y_{v2},\ldots,Y_{vn_v}\}$,
where $n_v$ is number of realizations (sample size). Getting one realization $Y_{vl}$ consists
of choosing value from each corresponding population $H_i,i\in I(v)$, and
calculation of $\phi_v$ value. By this we suppose that a random sample 
with replacement is used when each element from $H_i$ is choosen
independendly with the probability $1/n_i$. Further on, this procedure
is repeated for next vertex. Finally we get $Y_{k1},Y_{k2},\ldots,
Y_{kn_k}$ as values of the population $H_k$. Their average gives
the estimator $\Theta^*$ by analogy with formula (\ref{thetaest}).

\section{EXPRESSIONS FOR VARIANCE}

The aim of this section is to show how to calculate variance $D\;\Theta^*$ 
of the estimator (\ref{thetaest}).
It is easy to see that in the case of Hierarchical Bootstrap
the variance
$D\;\Theta^*$ is function of sample sizes $n_1, n_2,\ldots,n_k$.

In the previous papers of authors the variance $D\;\Theta^*$ was calculated
using the $\omega$-pairs notion (Andronov et. al, 1996, 1998). Then, it was
considered as continuos function of variables $n_1,n_2,\ldots,n_k$,
and reduced gradient method was used. But now we need other approach
for the calculation of $D\;\Theta^*$.

We use Taylor decomposition of function $\phi({\bf x})$ in respect to
mean ${\bf \mu}=(E\;X_1,E\;X_2,\ldots,E\;X_m)$:
\begin{equation}
\phi({\bf x})=\phi({\bf \mu})+\bigtriangledown^T\phi({\bf \mu})({\bf x}-{\bf \mu})+\frac{1}{2}({\bf x}-{\bf \mu})^T
\bigtriangledown^2\phi({\bf \mu})({\bf x}-{\bf \mu})+O(||{\bf x}-{\bf \mu}||^3),
\label{expr1}
\end{equation}
\begin{tabbing}
where \= $\bigtriangledown^2\phi({\bf x})$ is the matrix of second derivatives of
the function $\phi({\bf x})$,\\
\> $||{\bf v}||$ is Euclidean norm of vector ${\bf v}$.
\end{tabbing}

It gives the following decomposition:
\begin{eqnarray}
\phi({\bf x})\phi({\bf x'})=&\phi^2({\bf \mu})+\phi({\bf \mu})\bigtriangledown^T\phi({\bf \mu})({\bf x}-{\bf \mu})+
\phi({\bf \mu})\bigtriangledown^T\phi({\bf \mu})({\bf x'}-{\bf \mu})+\nonumber\\
&+\frac{1}{2}\phi({\bf \mu})({\bf x}-{\bf \mu})^T
\bigtriangledown^2\phi({\bf \mu})({\bf x}-{\bf \mu})+\nonumber\\
&+\frac{1}{2}\phi({\bf \mu})({\bf x'}-{\bf \mu})^T\bigtriangledown^2
\phi({\bf \mu})({\bf x'}-{\bf \mu})+\\
&+\bigtriangledown^T\phi({\bf \mu})({\bf x}-{\bf \mu})({\bf x'}-{\bf \mu})^T\bigtriangledown\phi({\bf \mu})+\nonumber\\
&+O(||{\bf x'}-{\bf \mu}||^3 + ||{\bf x}-{\bf \mu}||^3).\nonumber
\label{expr2}
\end{eqnarray}

If ${\bf X}=(X_1,X_2,\ldots,X_m)$ is a random vector with mutual independent components
$\{X_i\}$, $E\;{\bf X}={\bf \mu}$, $D\;X_i=\sigma^2_i$, then
\begin{equation}
E\;\phi({\bf X})=\phi({\bf \mu})+\frac{1}{2}\sum_{i=1}^{m}\sigma^2_i\frac{\partial^2}{\partial x_i^2}\phi({\bf \mu})+
E(O(||{\bf X}-{\bf \mu}||^3)),
\label{expr31}
\end{equation}

\begin{equation}
(E\;\phi({\bf X}))^2=\phi^2({\bf \mu})+\phi({\bf \mu})\sum_{i=1}^{m}\sigma_i^2\frac{\partial^2}{\partial x_i^2}\phi({\bf \mu})+\ldots.
\label{expr32}
\end{equation}

Now we suppose that $X_i$ and $X'_i$ are some values from sample population
$H_i=\{Y_{i1},Y_{i2},\ldots,Y_{in_i}\}$. Let $C_i$ denote the covariance
of two elements $Y_{il}$ and $Y_{il'}$ with different numbers $l$ and $l'$:
\begin{equation}
C_i=Cov(Y_{il},Y_{il'}|l\ne l').
\label{expr4}
\end{equation}

Let $X_i$ and $X'_i$ correspond to the elements $Y_{il}$ and $Y_{il'}$
accordingly. Because we extract $X_i$ and $X_i'$ from $H_i$ at random and
with replacement, then the event $\{l\ne l'\}$ occurs with the
probability $1/n_i$. Then
\begin{equation}
Cov(X_i,X_i')=\left\{
\begin{array}{ll}
Cov(Y_{il},Y_{il})=D\;Y_i&\mbox{with the probability $1/n_i$},\\
Cov(Y_{il},Y_{il'})=C_i&\mbox{with the probability $1-1/n_i$}.
\end{array}
\right.
\label{expr5}
\end{equation}
Therefore
\begin{equation}
Cov(X_i,X_i')=\frac{1}{n_i}D\;Y_i+\left(1-\frac{1}{n_i}\right)C_i.
\label{expr6}
\end{equation}

If the values $\phi(X)$ and $\phi(X')$ correspond to subfunction $y_v=\phi_v(\cdot)$
and the sample population $H_v$, then formulas (\ref{expr2}), (\ref{expr31}) and (\ref{expr32})
give
\begin{equation}
C_v=\sum_{i\in I_v}\left(\frac{\partial}{\partial x_i}\phi_v(\mu_v)\right)^2 Cov(X_i,X_i')+\ldots.
\label{expr7}
\end{equation}

Now we can get from (\ref{expr6})
\begin{equation}
Cov(\phi_v(X),\phi_v(X'))=\frac{1}{n_v}\sigma^2_v+\left(1-\frac{1}{n_v}\right)C_v,
\label{expr8}
\end{equation}
where the variance $\sigma^2_v=D\;Y_v$ can be determined from (\ref{expr7}) by $X_i=X_i'$:
\begin{equation}
\sigma^2_v=\sum_{i\in I_v}\left(\frac{\partial}{\partial x_i}\phi_v(\mu_v)\right)^2\sigma_i^2+\ldots.
\label{expr9}
\end{equation}

Finally we have
\begin{equation}
Cov(\phi_v(X),\phi_v(X'))=\sum_{i\in I_v}\left(\frac{\partial}{\partial x_i}
\phi_v(\mu_v)\right)^2\left[\frac{1}{n_v}\sigma_i^2+\left(1-\frac{1}{n_v}\right)Cov(X_i,X_i')\right]+\ldots,
\label{expr10}
\end{equation}
or
$$
Cov(\phi_v(X),\phi_v(X'))=\sum_{i\in I_v}\left(\frac{\partial}{\partial x_i}
\phi_v(\mu_v)\right)^2\left[
\left(\frac{1}{n_v}+\left(1-\frac{1}{n_v}\right)\frac{1}{n_i}\right)\sigma^2_i+\ldots\right.
$$
\begin{equation}
\left.+\left(1-\frac{1}{n_v}\right)\left(1-\frac{1}{n_i}\right)C_i\right].
\label{expr11}
\end{equation}

By this we suppose that variances  $\sigma_1^2,\sigma_2^2,\ldots,
\sigma_m^2$ of input random variables $X_1,X_2,\ldots,X_m$ are
known. For example, it is possible to use estimators of these variances,
calculated on given sample populations $H_1,H_2,\ldots,H_m$.

If the vertex $i$ belongs to the initial level of the calculation
tree (it means that $i=1,2,\ldots,m$) then $\phi_i(x_i)=x_i$,
$\sigma_i^2$ is known value, $C_i=0$. Therefore
\begin{equation}
Cov(\phi_i(X_i),\phi_i(X'_i))=\frac{1}{n_i}\sigma_i^2,\qquad i=1,2,\ldots,m.
\label{expr12}
\end{equation}

Another covariances and variances are calculated recurrently in accordance
with formulas (\ref{expr7}),(\ref{expr9}), (\ref{expr11}), (\ref{expr12})
from vertices with less numbers to vertices with great numbers. Finally
we get the variance $D\;\Theta^*$ of interest as the variance for root of the
calculation tree:
\begin{equation}
D\;\Theta^*=\frac{1}{r}(\sigma_k^2+(r-1)C_k)+\ldots,
\label{expr13}
\end{equation}
where $r=n_k$.

As it was just mentioned, we will consider the variance $D\;\Theta^*$ as
a function of sample sizes $n_1,n_2,\ldots,n_k$ and denote it
$D(n_1,n_2,\ldots,n_k)$. Our aim is to minimize this function in respect to variables
$n_1,n_2,\ldots,n_k$ by linear restriction, or, by other words, to solve
the following optimization problem:
\begin{equation}
\mbox{minimize }D(n_1,n_2,\ldots,n_k)
\label{probl}
\end{equation}
by restriction
\begin{equation}
a_1n_1+a_2n_2+\ldots+a_kn_k\le b,
\label{restric}
\end{equation}
where $b$, $\{a_i\}$ and $\{n_i\}$ are integer non-negative numbers.

Now we intent to apply the dynamic programming method (Minox 1989).

\section{THE DYNAMIC PROGRAMMING METHOD}

Let us
solve the optimization problem (\ref{probl}), (\ref{restric}). 
Our function of interest $\phi({\bf x})$ is calculated and simulated
recurrently, using the calculation tree (see Section 2). 
In accordance to the dynamic programming technique, we have
"forward" and "backward" procedure.

During "backward" procedure, we calculate recurrently so-called Bellman
function $\Phi_v(\alpha,z)$, $v=1,2,\ldots,k$, $z=1,2,\ldots,b$, $0\le\alpha\le 1$.
Let us consider the subfunction $\phi_v(\cdot)$, that corresponds to the
vertex $v$. This subfunction directly depends on variables $y_i$, $i\in I_v$,
which correspond to incoming arcs for the vertex $v$. Additionally $\phi_v$
depends on variables $\{x_i\}$ and $\{y_i\}$ from which there exists path from leaves to the
vertex $v$ of our calculation tree. Let $B_v$ denote corresponding set
of variables $\{x_i\}$ and $\{y_i\}$.

Now we need to denote some auxiliary functions. Let us introduce the following
notation
\begin{equation}
\psi_i(\alpha)=\alpha\sigma_i^2+(1-\alpha)Cov(X_i,X'_i),\qquad i=1,2,\ldots,k,\;0\le\alpha\le 1.
\label{dyn1}
\end{equation}

Then we are able to write in accordance with (\ref{expr10}):
\begin{equation}
Cov(\phi_v(X),\phi_v(X'))=\sum_{i\in I_v}\left(\frac{\partial}{\partial x_i}\phi_v(\mu_v)\right)^2\psi_i\left(\frac{1}{n_v}\right).
\label{dyn2}
\end{equation}

Now we have from (\ref{dyn1}), (\ref{expr9}) and (\ref{expr10})
$$
\psi_v(\alpha)=\sum_{i\in I_v}\left(\frac{\partial}{\partial x_i}\phi_v(\mu_v)\right)^2
\left\{\alpha\sigma_i^2+(1-\alpha)\left[\frac{1}{n_v}\sigma_i^2+
(1-\frac{1}{n_v})Cov(X_i,X_i')\right]\right\}=
$$
$$
\sum_{i\in I_v}\left(\frac{\partial}{\partial x_i}\phi_v(\mu_v)\right)^2
\left[\left(\alpha+\frac{1-\alpha}{n_v}\right)\sigma_i^2+(1-\alpha)(1-\frac{1}{n_v})Cov(X_i,X'_i)\right],
$$
so
\begin{equation}
\psi_v(\alpha)=\sum_{i\in I_v}\left(\frac{\partial}{\partial x_i}\phi_v(\mu_v)\right)^2
\psi_i\left(\alpha+\frac{1-\alpha}{n_v}\right).
\label{dyn3}
\end{equation}

Note that it follows from (\ref{expr8}) and (\ref{dyn1}) that our variance
of interest (\ref{expr13}) is
\begin{equation}
D\;\Theta^*=\psi_k(0).
\end{equation}

Values $\psi_v(\alpha)$ depend on the sample sizes $n_i$ for all $i\in B_v$.
We will mark this fact as $\psi_v(\alpha)=\psi_v(\alpha;n_i:i\in B_v)$.

Now we are able to introduce above mentioned Bellman functions:
\begin{equation}
\Phi_v(\alpha,z)=\min_{n_i}\psi_v(\alpha;n_i:i\in B_v),
\label{dyn5}
\end{equation}
where minimization is realized with respect to non-negative integer
variables $n_i$ that are satisfied the linear restriction
\begin{equation}
\sum_{i\in B_v}a_in_i\le z.
\label{dyn6}
\end{equation}

It is clear that optimal value of variance $D^*(\cdot)$ for the problem
(\ref{probl}), (\ref{restric}) is equal to $\Phi_k(0,b)$.

Bellman functions $\Phi_v(\alpha,z)$ are calculated recurrently
from $v=1$ to $v=k$ for $0\le\alpha\le 1$ and $z=1,2,\ldots,b$.
Basic functional equation of dynamic programming has the
following form:
\begin{equation}
\Phi_v(\alpha,z)=\min\sum_{i\in I_v}\left(\frac{\partial}{\partial x_i}\phi_v(\mu_v)\right)^2
\Phi_i\left(\alpha+\frac{1-\alpha}{n_v};z_i\right)
\label{dyn7}
\end{equation}
where minimaiztion is realized with respect to non-negative
integer variables $n_v$ and $\{z_i:i\in I_v\}$ that satisfy the
linear restriction
\begin{equation}
a_vn_v+\sum_{i\in I_v}z_i\le z.
\label{dyn8}
\end{equation}

The initial values of $\Phi_v(\cdot)$ are determined with the tree leaves
by formulas (\ref{expr12}), (\ref{dyn1}) and (\ref{dyn5}):
\begin{equation}
\Phi_v(\alpha,z)=\alpha\sigma^2_v+(1-\alpha)\frac{1}{[z/a_v]}\sigma_v^2=
\label{dyn9}
\end{equation}
$$
=\sigma_v^2\left(\alpha+(1-\alpha)\frac{1}{[z/a_v]}\right),\qquad v=1,2,\ldots,m,
$$
where $[z/a_v]$ - integer part of number $z/a_v$,

Thus the "backward" procedure is a recurrent calculation of Bellman
functions $\Phi_v(\alpha,z)$ for $v=1,2,\ldots,k$, $0\le\alpha\le 1$, $z=1,2,\ldots,b$
by using formulas (\ref{dyn9}), (\ref{dyn7}). Finally we get the minimal
variance
\begin{equation}
D^*\;\Theta^*=\Phi_k(0,b).
\label{dyn10}
\end{equation}

To calculate the optimal sample sizes $n^*_1,n^*_2,\ldots,n^*_k$ we should
apply "forward" procedure of dynamic programming technique.
At first, we find $n^*_k$ and $\{z^*_i:i\in I_k\}$ by solving the
equation
\begin{equation}
\Phi_k(0,b)=\min\sum_{i\in I_k}\left(\frac{\partial}{\partial x_i}\phi_k(\mu_k)\right)^2
\Phi_i\left(\frac{1}{n^*_k},z_i^*\right)
\label{dyn11}
\end{equation}
where minimization is realized by condition
\begin{equation}
a_kn_k^*+\sum_{i\in I_k}z_i^*\le b.
\label{dyn12}
\end{equation}

Let $I^{-1}(v)=\{\omega:v\in I(\omega)\}$, $\alpha_k=1/n_k^*$.
Then we recurrently determine by analogy the rest $n_v^*$ and $\{z^*_i:i\in I_v\}$
for $v=k-1,k-2,\ldots,m+1$:
\begin{equation}
\Phi_v(\alpha_{I^{-1}(v)},z^*_v)=\min\sum_{i\in I_v}\left(\frac{\partial}{\partial x_i}\phi_v(\mu_v)\right)^2
\Phi_i\left(\alpha_{I^{-1}(v)}+\frac{1-\alpha_{I^{-1}(v)}}{n_v^*},z_i^*\right)
\label{dyn13}
\end{equation}
by condition
\begin{equation}
a_vn_v^*+\sum_{i\in I_v}z_i^*\le z_v^*.
\label{dyn14}
\end{equation}

Moreover we put
\begin{equation}
\alpha_v=\alpha_{I^{-1}(v)}+(1-\alpha_{I^{-1}(v)})/n_v^*
\label{dyn15}
\end{equation}

Finally the optimal sizes $n_i^*$ for $i=1,2,\ldots,m$ are determined by the
following way:
\begin{equation}
n_i^*=[z_i^*/a_i].
\label{dyn16}
\end{equation}

\section*{References}
\begin{description}
\item 
Andronov, A., Merkuryev,~Yu.~(1996) Optimization of Statistical Sizes
in Simulation. In: {\sl Proceedings
of the 2-nd St. Petersburg Workshop on Simulation}, St. Petersburg
State University, St. Petersburg, 220-225.
\item
Andronov,~A., Merkuryev,~Yu.~(1998) Controlled Bootstrap Method and
its Application in Simulation of Hierarchical Structures. In: {\sl Proceedings
of the 3-d St. Petersburg Workshop on Simulation}, St. Petersburg
State University, St. Petersburg, 271-277.
\item
Davison,~A.C., Hinkley,~D.V.~(1997) {\sl Bootstrap Methods and their
Application}. Cambridge university Press, Cambridge.
\item
Efron,~B., Tibshirani,~R.Y.~(1993) {\sl Introduction to the Bootstrap}.
Chapman \& Hall, London.
\item
Minox,~M.~(1989) {\sl Programmation Mathematique. Teorie et Algorithmes.}
Dunod.

\end{description}

\end{document}